\documentclass{article}
\usepackage{ijcai20}
\usepackage{times}
\usepackage{soul}
\usepackage{url}
\usepackage[backref]{hyperref}
\usepackage[utf8]{inputenc}
\usepackage[small]{caption}
\usepackage{graphicx}
\usepackage{amsthm}
\usepackage{booktabs}
\usepackage{algorithm}
\usepackage{amsmath,amssymb}
\usepackage[american]{babel}
\usepackage{array}
\usepackage{bm}
\usepackage{subfigure}
\usepackage{algorithmicx}
\usepackage{algpseudocode}
\newcommand{\x}{{\bm x}}
\newcommand{\z}{{\bm z}}
\newcommand{\p}{{\bm c}}
\newcommand{\q}{{\bm q}}
\newcommand{\h}{{\bm h}}

\newcommand{\T}{{\top}}
\newcommand{\argsort}{\mathop{\rm arg~sort}\limits}
\newcommand{\argmax}{\mathop{\rm arg~max}\limits}

\title{Mutual Teaching for Graph Convolutional Networks}
\author{
Kun Zhan, Chaoxi Niu
    \affiliations
    School of Information Science \& Engineering, Lanzhou University, China
    \emails
    kzhan@lzu.edu.cn
}
\begin{document}
\maketitle
\begin{abstract}
Graph convolutional networks produce good predictions of unlabeled samples due to its transductive label propagation. Since samples have different predicted confidences, we take high-confidence predictions as pseudo labels to expand the label set so that more samples are selected for updating models. We propose a new training method named as mutual teaching, i.e., we train dual models and let them teach each other during each batch. First, each network feeds forward all samples and selects samples with high-confidence predictions. Second, each model is updated by samples selected by its peer network. We view the high-confidence predictions as useful knowledge, and the useful knowledge of one network teaches the peer network with model updating in each batch. In mutual teaching, the pseudo-label set of a network is from its peer network. Since we use the new strategy of network training, performance improves significantly. Extensive experimental results demonstrate that our method achieves superior performance over state-of-the-art methods under very low label rates.
\end{abstract}

\section{Introduction}
Although graph convolutional network (GCN)~\cite{kipf2016semi} recently made great achievement in semi-supervised learning (SSL) and many GCN-based SSL algorithms were developed, GCN-based SSL methods currently still have some issues:

How to expand the label set. To expand the label set, we utilize predictions of GCN to expand the label set to select more samples involved in updating models. Since GCN has the property of transductive label propagation, i.e., Laplacian smoothing \cite{li2018deeper}, each connected component in a graph tends to have the same label. Labeled nodes propagate its label to unlabeled nodes in the graph. We select samples with high-confidence prediction probabilities produced by the {\rm softmax} layer of GCN as pseudo labels.

How to improve the training strategy. To further improve the network performance, we use dual models with a new training strategy called as mutual teaching. In each batch, each network regards its high-confidence predictions as the useful knowledge and teaches the knowledge to its peer network. Both soft and hard predicted targets are exploited to improve the performance. One network generates useful knowledge and the other learns from its peer network.

Most of them require many labeled data. We consider SSL when the labeled samples are very limited, i.e., even two or three labeled samples per class are available for training a model, which is a very challenge problem. Most of the existing GCN-based SSL methods used 20 labeled samples per class to train only one GCN model, and they did not learn the models when very few labeled samples are available. Compared with these algorithms, SSL from very few labeled data is vitally important. Due to the expensive labeling cost, it is hard to obtain many labeled data.

\begin{figure*}[ht]
\centering
\includegraphics[width=0.8\textwidth]{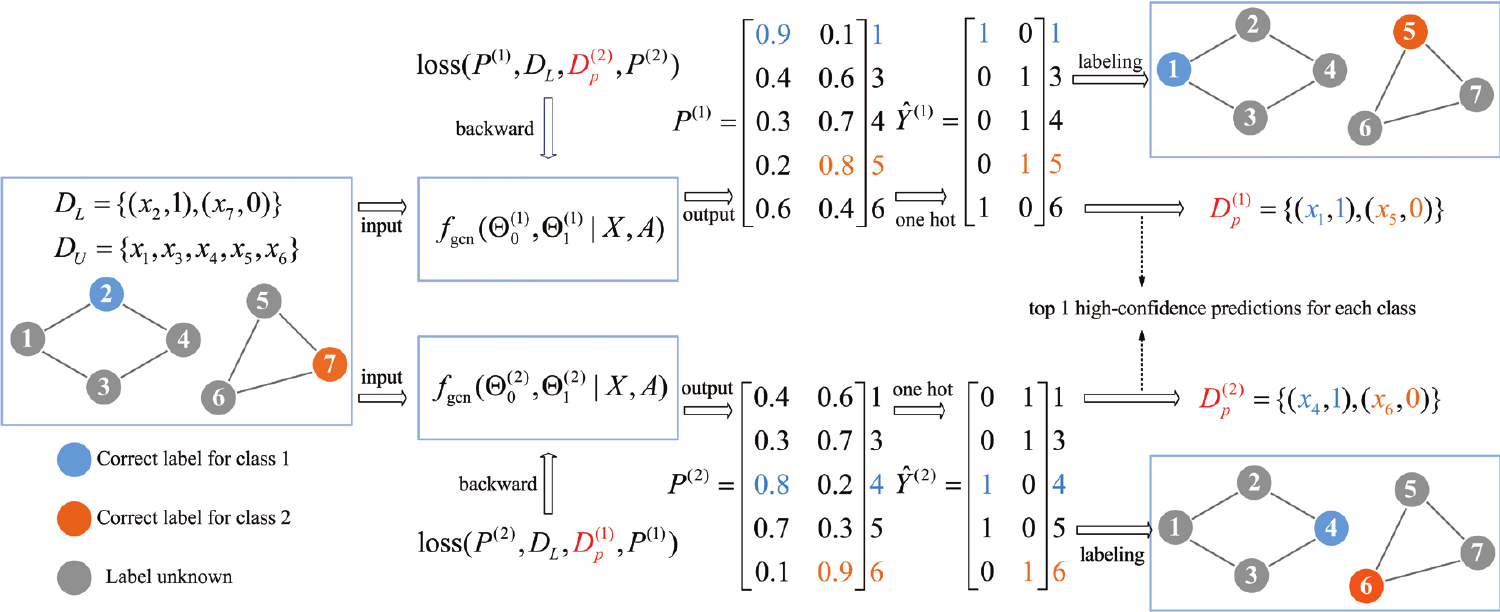}
\caption{A simple example of mutual teaching algorithm in dual models.}\label{MVGL2}
\end{figure*}

To address these issues, we propose mutual teaching for GCN (MT-GCN). We present a new MT-GCN SSL algorithm to overcome the limits mentioned above. Fig.~\ref{MVGL2} shows a simple example of mutual teaching algorithm in dual models. This method trains dual GCN models and then learns from each other with the most confidence pseudo labels. What is more, each model is updated with three loss terms, a supervised loss with labeled samples, a pseudo-label loss, and a consistency loss. Except for the supervised loss, the two other loss terms use the mutual teaching strategy. The pseudo-label loss uses hard pseudo targets while the consistency loss uses soft targets produced by the softmax layer of GCN. Overall, the contributions of MT-GCN are summarized below:

1) We use dual GCN models for improving the prediction performance. With very small number of labeled data (e.g., even two or three samples are available per class) and the exploited pseudo labels, the performance of the proposed MT-GCN is better than other GCN-based algorithms in SSL.

2) Different from only using the loss function with labeled samples, we use the pseudo labels to calculate two new loss terms. The new loss terms are designed to encourage implicitly cross-model prediction alignment for each class from both labeled samples and selected pseudo labels.

3) We obtain high quantitative metrics, especially when only two or three labeled samples per class are available. Extensive experiments demonstrate that our method is better than state-of-the-art approaches in all considered datasets with a small number of labeled data.

The rest of the paper is organized as follows: Section \ref{RW} introduces some related work. GCN are introduced in Section \ref{Preliminaries}. In Section \ref{sec_MLGCN}, we propose our MT-GCN method to solve SSL with very few labeled data. In Section \ref{Experiments}, we conduct experiments to demonstrate the effectiveness of our proposed method. In Section \ref{Conclusions}, we present conclusion of the paper.
\section{Related Work}\label{RW}
\subsection{Graph Neural Networks}
Graph neural networks are widely exploited for machine learning tasks recently~\cite{Wu9046288,ji2020survey,Pan8822591}. GCN~\cite{kipf2016semi} was applied to SSL because the graph convolution of GCN is a special form of Laplacian smoothing over the graph~\cite{li2018deeper}. GraphSAGE~\cite{Hamilton2017} proposed an inductive framework that generates embedding by sampling and aggregating features from a node's local neighborhood. GAT~\cite{velivckovic2017graph} applied the multi-head self-attention mechanism to parameterize the edge weight. Xu {\it et al.},~\cite{xu2018powerful} discussed the representational capacity by analyzing different structures and proposed a graph isomorphism network. Wu {\it et al.}~\cite{wu2020unsupervised} used GCN for unsupervised domain adaption.
\subsection{Self-Supervised Learning}
Self-supervised learning utilizes auxiliary tasks to produce pseudo labels~\cite{doersch2015unsupervised}. Most self-supervised learning algorithms used the same network architecture in both the pretext task and the fine-tuning task~\cite{goyal2019scaling}. For transferring knowledge between two models, knowledge distillation~\cite{bu2006model,hinton2015distilling} can use to transfer the representation in a trained model with the pretext task to the other one employed on the target task. DeepCluster~\cite{caron2018deep} took a set of embedding features and grouped them into different clusters to generate pseudo labels. In GCN-based SSL, Li {\it et al.},~\cite{li2018deeper} proposed two strategies to train GCN with very few labeled data and showed GCN might result in features within connected component converging to the same value. Sun {\it et al.},~\cite{sun2019multi} also used very few labeled data and utilized pseudo labels to expand the label set. Two methods, MultiStage and M3S, were proposed. In MultiStage, it added most confidence vertices with predicted pseudo labels to expand the label set in each stage. M3S additionally utilized algorithm self-checking mechanism to choose nodes with precise pseudo labels.
\subsection{Learning with Dual Models}
Knowledge distillation~\cite{hinton2015distilling} is mainly used to transfer the good performance of a large model to a small model. Inspired by knowledge distillation, Zhang {\it et al.},~\cite{zhang2018deep} employed two convolutional neural networks with Kullback-Leibler divergence (KLD) for alignment the feature maps. For exploiting complementary of two models, Wu {\it et al.},~\cite{wu2019mutual} designed a subnet to capture complementary information for image classification. Mean Teacher~\cite{tarvainen2017mean} improved performance of Student by a knowledge distillation loss between Student and Teacher while the input of Student is degraded by noise, and Teacher copied the averaged weights of Students. Different from Teacher-Student strategy, we use a mutual teaching strategy. In knowledge distillation~\cite{hinton2015distilling} and label smoothing~\cite{muller2019does}, both of them exploit hard and soft targets. Knowledge distillation directly changes the temperature of softmax layer, while label smoothing directly changes the ground-truth labels. In MT-GCN, both hard and soft targets are used for different purposes, hard targets are used to expand the label set while soft targets are used to each network matches its peer network.
\section{Graph convolutional networks}\label{Preliminaries}
$G=(V,E)\,$ denotes a graph, its vertex set is denoted by $V$, and its edge set is $E$. There are $|V|=n$ vertices in $G$. Its edge is described by an affinity matrix $A=[a_{ij}]\in \mathbb{R}^{n \times n}$ and $a_{ij}$ denotes the pairwise connection weight between two vertices. Each vertex corresponds to a data vector $\x$ and the data matrix $X=[\x_1,\x_2,\ldots,\x_n]^\T \in\mathbb{R}^{n \times c}$ has $n$ data points and $c$ input channels.

In the graph Fourier domain~\cite{shuman2013emerging}, Fourier coefficients $\hat{\x}$ is transformed by a spatial domain signal $\x$, i.e., $\hat{\x}=U^\T \x\,$, where $U$ is a Fourier basis. The inverse transform is $\x=U \hat{\x}\,$. the Fourier basis of GCN is a matrix of eigenvectors of the normalized Laplacian $L=I_n-D^{-\frac{1}{2}}AD^{-\frac{1}{2}}=U\Lambda U^\T\,$, where $\Lambda$ is a diagonal matrix of eigenvalues of $L\,$, $D$ is the degree matrix of $A$, i.e., $d_{ii}=\sum_ja_{ij}$, and $I_n$ denotes an identity matrix.

According to convolution theorem, $x$ convolutes a filter $\h$ is given by,
\begin{equation}\label{ct}
\x\otimes \h=U(\hat{\h}\odot\hat{\x})=U{\rm diag}(\hat{\h})U^\T\x
\end{equation}
where $\hat{\h}=U^\T\h\,$ is a vector of Fourier coefficients of a filter $\h$, $\otimes$ denotes the graph convolution operator, and $\odot$ is the element-wise Hadamard product.

$K$-th order Chebyshev approximation of ${\rm diag}(\hat{\h})$~\cite{hammond2011wavelets} is given by,
\begin{equation}\label{cp}
{\rm diag}(\hat{\h})\approx\sum_{{i}=0}^{K}\theta_{i}\left(\frac{ 2\Lambda}{\lambda_{\max}}-I_n\right)^{i}
\end{equation}
where $\theta_{i}$ is the polynomial coefficient and $\lambda_{\max}$ is the largest eigenvalue of $L$.

Since $(U\Lambda U^\T)^{i}=U\Lambda^{i} U^\T$, we substitute Eq.~\eqref{cp} into Eq.~\eqref{ct},
\begin{equation}\label{ct3}
\x\otimes \h\approx\sum_{{i}=0}^{K}\theta_{i} \left(\frac{ 2L}{\lambda_{\max}}-I_n\right)^{i}\x\,.
\end{equation}

Employing a localized first-order truncated Chebyshev polynomial approximation~\cite{defferrard2016convolutional,kipf2016semi}, Eq.~\eqref{ct3} simplifies to,
\begin{equation}\label{ct4}
\x\otimes \h\approx\theta(I_n+D^{-\frac{1}{2}}AD^{-\frac{1}{2}})\x\,.
\end{equation}

Kipf and Welling~\cite{kipf2016semi} renormalized $I_n+D^{-\frac{1}{2}}AD^{-\frac{1}{2}}$ to $\hat{A}=\tilde{D}^{-\frac{1}{2}}\tilde{A}\tilde{D}^{-\frac{1}{2}}$ with $\tilde{A}=A+I_n$ and $\tilde{d}_{ii}=\sum_j\tilde{a}_{ij}\,$.

Thus, given the graph $\hat{A}$ and the matrix $X$, a prorogation layer of GCN is defined by,
\begin{equation}\label{layer}
Z = {\rm ReLU}(\hat{A}X\Theta)
\end{equation}
where $\Theta\in \mathbb{R}^{c\times f}$ has $f$ number of filters, $Z$ is the convolved feature matrix, and ${\rm ReLU}(\cdot) = \max(0, \cdot)\,$ is the nonlinear activation function.

In this paper, we use a two-layer GCN model according to~\cite{kipf2016semi},
\begin{equation}\label{gcn1}
Z = \hat{A}~{\rm ReLU}(\hat{A}X\Theta_0)\Theta_1
\end{equation}
where the output feature map $Z=[z_{ij}]\in\mathbb{R}^{n\times k}$ is the logit.

The prediction probability of a sample $x_i$ given by the GCN model is computed as
\begin{eqnarray}
{\bm p}_i&=&{\rm softmax}(\z_i)\\
& =& \frac{\exp(\z_i)}{\sum_j\exp(z_{ij})}
\end{eqnarray}
where ${\bm p}_i$ is a row vector and the {\rm softmax} output of the GCN model.
\section{Mutual teaching GCN}\label{sec_MLGCN}
Suppose that there are $k$ classes in the data matrix $X$. The labeled data is denoted by $D_L=\{(\x_i, y_{ij}), \forall~i\in V_L,j\in[1,k]\}$ and the unlabeled data is $D_U=\{\x_i, \forall~i\in V_U\}$, where $V_L$ is the labeled vertices set, $V_U$ is the unlabeled set, and $V = V_L \cup V_U$.

The goal of SSL is to exploit labeled and unlabeled data to predict the label of the data in $V_U$ when the number of labeled data is very few.
\subsection{Top $t$ High-Confidence Predictions}\label{topt}
As shown in Fig.~\ref{MVGL2}, pseudo labels $\hat{y}_{ij}$ of $i\in D_U$ easily obtains from $p_{ij}$ by using a one-hot operation,
\begin{equation}\label{yij}
\begin{cases}
j=\argmax([p_{i1},p_{i2},\ldots,p_{ik}]),\\
\hat{y}_{ij}=1
\end{cases}
\end{equation}
where the column index $j$ of the maximum of a row $i$ is the pseudo label of the data point $\x_i$ and then the $(i,j)$-th element of $\hat{Y}$ is set to 1.

The prediction confidence $c_i$ of a data point $\x_i$ is assigned to the maximum of the $i$-th row $\bm{p}_i$ of the matrix $P$,
\begin{equation}\label{pc}
c_i=\max([p_{i1},p_{i2},\ldots,p_{ik}])\,.
\end{equation}

Then, we sort elements in the column vector $\p=[c_1;c_2;\ldots;c_n]$ in descending order, and it returns the index set,
\begin{equation}\label{q_index}
\q=\argsort(\p)=\argsort\left(
\left[
  \begin{array}{c}
    c_1 \\
    c_2 \\
    \vdots \\
    c_n \\
  \end{array}
\right]
\right)
\end{equation}
where $\q$ returns an index vector of the ordered confidence.

With the index vector of the ordered confidence, it is easy to obtain top $t$ pseudo labels for each class. Specifically, for each single GCN, i.e., the $g$-th GCN $\forall\,g\in\{1,2\}$, we use $\q$ to obtain its index set $V^{(g)}$ of the top $t$ pseudo labels for each class as shown in Algorithm~\ref{alg1}.
\begin{algorithm}[!htbp]
\caption{Top $t$ high-confidence predictions for each class.}\label{alg1}
\begin{algorithmic}[1]
\State {\bf Input:} $n,k,t,D_U,\hat{y}_{ij}$ and $\q$\,.
\State {\bf Output:} $V^{(g)}=\{idx_i,idx_2,\ldots,idx_{t\times k}\}$.
\State {\bf Initialize:} $cnt_j\leftarrow0,\forall~j\in[1,k]$ and $m\leftarrow1$\,.
\For{$i\in\{q_1,q_2,\ldots,q_n\}$}
    \For{$j\in[1,k]$}

        \If{$cnt_j<=t$\,\&\,$\hat{y}_{ij}=1$\,\&\,$i\in D_U$}
            \State $idx_m\leftarrow i$\,.
            \State $m\leftarrow m+1$\,.
            \State $cnt_j\leftarrow cnt_j+1$\,.
        \EndIf
    \EndFor
\EndFor
\end{algorithmic}
\end{algorithm}
\subsection{Supervised Loss}
For semi-supervised multi-class classification, Kipf and Welling~\cite{kipf2016semi} evaluated the cross-entropy loss over the labeled data set $D_L$,
\begin{equation}\label{loss_gcn}
\mathcal{L}_{\rm sup}=-\sum_{i\in V_L}\sum_{j=1}^k y_{ij}\ln p_{ij}
\end{equation}
where $p_{ij}$ is the $(i,j)$-th element of $P$.

In this paper, we call Eq.~\eqref{loss_gcn} as the supervised loss function since it only uses the labeled samples.
\subsection{Pseudo-label Loss}
Besides the supervised loss, Eq.~\eqref{loss_gcn}, we exploit two other loss functions, a pseudo-label loss and a consistency loss. Both the two loss functions use pseudo labels for mutual teaching.

For two GCN models, we obtain two index sets $V^{(1)}$ and $V^{(2)}$ by using Algorithm~\ref{alg1}, respectively. After we obtain the top $t$ high-confidence predictions for each class, we can expand them to the label set for updating model. Besides adding a loss function with the pseudo labels directly, we use information entropy as a measure of uncertainty~\cite{iscen2019label} to assign a weighted value for each sample $\x_i$. Given a probability $\bm{p}_i$ of a sample $\x_i$, its certainty $w_i$ can be defined by,
\begin{equation}\label{certainty}
w_i=1-\frac{{\rm H}(\bm{p}_i)}{\log k}
\end{equation}
where ${\rm H}(\cdot)$ is the information entropy.

Eq.~\eqref{certainty} shows it tends to zero if all of elements in $\bm{p}_i$ are $\frac{1}{k}$ and it is assigned to a high value if $\bm{p}_i$ is one-hot.

Then, the pseudo-label loss of the first model is defined by,
\begin{equation}\label{loss_pl1}
\mathcal{L}_{\rm pl}^{(1)}=-\frac{1}{|V^{(2)}|}\sum_{i\in V^{(2)}}w_i\sum_{j=1}^k \hat{y}_{ij}^{(2)}\log p^{(1)}_{ij}\,.
\end{equation}

Similarly, the pseudo-label loss of the second model is given by,
\begin{equation}\label{loss_pl2}
\mathcal{L}_{\rm pl}^{(2)}=-\frac{1}{|V^{(1)}|}\sum_{i\in V^{(1)}}w_i\sum_{j=1}^k \hat{y}_{ij}^{(1)}\log p^{(2)}_{ij}\,.
\end{equation}
\subsection{Consistency Loss}
The consistency loss function encourages consistency under different network embedding of the same data in each batch. To quantify the prediction consistency of the dual GCN models, we use KLD as the consistency loss.

KLD from ${\bm p}^{(1)}_i$ and $\bm{p}^{(2)}_i$ is given by,
\begin{equation}\label{loss_cl1}
\mathcal{L}_{\rm cl}^{(1)}=\sum_{i\in V^{(2)}}\sum_{j=1}^k p^{(2)}_{ij}\log\frac{p^{(2)}_{ij}}{p^{(1)}_{ij}}\,.
\end{equation}

KLD from ${\bm p}^{(2)}_i$ and $\bm{p}^{(1)}_i$ is given by,
\begin{equation}\label{loss_cl1}
\mathcal{L}_{\rm cl}^{(2)}=\sum_{i\in V^{(1)}}\sum_{j=1}^k p^{(1)}_{ij}\log\frac{p^{(1)}_{ij}}{p^{(2)}_{ij}}\,.
\end{equation}

While each one learns to match the probability of its peer with consistency loss functions, each network learns to correctly predict true labels with the supervised loss Eq.~\eqref{loss_gcn},
\subsection{Mutual Teaching}
Mutual teaching approach is formulated by a cohort of dual GCN models.

In section \ref{topt}, we mainly attain two index sets $V^{(1)}$ and $V^{(2)}$ of the top $t$ high-confidence predictions to expand the label set. As shown in Fig.~\ref{MVGL2}, it is straightforward to check that pseudo-label set can  $V^{(1)}$ and $V^{(2)}$ easily obtain.

In the two loss functions, one network uses an index set from its peer network. The high-confidence predictions of one network teach its peer network to update its model. In mutual teaching, the pseudo-label loss uses hard targets $\hat{y}_{ij}$ while the consistency loss uses soft targets $p_{ij}$. Cross-updating dual networks, one network learns knowledge from the peer network. Hard targets mainly expand the label set while the soft targets improve model calibration, which can significantly improve performance.

The overall loss function of each model is given by,
\begin{equation}\label{Ols}
\mathcal{L}_{o}^{(g)}=\mathcal{L}_{\rm sup}^{(g)}+\mathcal{L}_{\rm pl}^{(g)}+ \mathcal{L}_{\rm cl}^{(g)},  \forall~g\in\{1,2\}\,.
\end{equation}

The detailed algorithm of the proposed method is summarized in Algorithm~\ref{alg2}.
\begin{algorithm}[ht]
\caption{Mutual teaching GCNs}\label{alg2}
\begin{algorithmic}[1]
\State {\bf Input:} $D_L$, $D_U$, $A$, $N$.
\State {\bf Output:} $Z^{(1)}$ and $Z^{(2)}\,$.
\State {\bf Initialize:} $\hat{A}=\tilde{D}^{-\frac{1}{2}}\tilde{A}\tilde{D}^{-\frac{1}{2}}$ with $\tilde{A}=A+I_n$ and $\tilde{D}_{ii}=\sum_j\tilde{A}_{ij}\,$. $\Theta_0^{(1)}$, $\Theta_1^{(1)}$, $\Theta_0^{(2)}$, and $\Theta_1^{(2)}$ from scratch.
\For{$epoch\in[1,N]$}
    \For{$g\in\{1,2\}$}
        \State Update $Z^{(g)}$ by Eq.~\eqref{gcn1}.
        \State Update $\hat{y}_{ij}^{(g)}$ by Eq.~\eqref{yij}.
        \State Update $\p^{(g)}$ by Eq.~\eqref{pc}.
        \State Update $\q^{(g)}$ by Eq.~\eqref{q_index}.
        \State Update $V^{(g)}$ by Algorithm~\ref{alg1}.
    \EndFor
    \For{$g\in\{1,2\}$}
        \State Update the model with  Eq.~\eqref{Ols}.
    \EndFor
\EndFor
\end{algorithmic}
\end{algorithm}

In mutual teaching, we exchange indices of pseudo-label sets. With the pseudo-label loss and the consistency loss, each network teaches knowledge to its peer network. The high-confidence predictions are useful knowledge, so it teaches its peer network with such knowledge.
\section{Experiments}\label{Experiments}
In this section, we conduct experiments on three popular benchmarks with different label rates to demonstrate the effectiveness of our proposed MT-GCN method. We evaluate the performance by the metric of classification accuracy.
\subsection{Datasets}
Three widely used  citation datasets are used in this paper:
\begin{itemize}
    \item {\bf Cora:} Cora consists of seven classes with 2708 scientific publications and contains 5429 citation links. Each publication is described by a bag-of-words feature, i.e., a 0/1 value vector indicates the absence/presence of a certain word. The feature dimension of a publication of Cora is 1433. We evaluate MT-GCN under different label rates, 0.5\%, 1\%, 2\%, and 3\%, i.e., 2, 4, 8, and 12 per class.

  \item {\bf Citeseer:} Citeseer contains 3327 scientific publications which are classified into six classes and has 4732 citation links. The feature dimension of a publication of Citeseer is 3703. We evaluate MT-GCN under different label rates over Citeseer: 0.5\%, 1\%, 2\%, and 3\%, i.e., 3, 6, 12, and 18 per class.

  \item {\bf PubMed:} PubMed consists of three classes with 19717 scientific publications and contains 44338 citation links. The feature dimension of a publication of PubMed is 500. We evaluate MT-GCN under different label rates: 0.03\%, 0.05\%, and 0.1\%, i.e., 2, 3, and 7 per class.
\end{itemize}

The statistics of these three datasets are summarized in Table~\ref{dataset}.
\begin{table}[!ht]
  \caption{Datasets statistics.}
  \label{dataset}
  \centering
  \begin{tabular}{l|cccc}
    \hline
    Dataset & Nodes & Edges & Classes & Dimensions \\
    \hline
    Cora & 2708 & 5429 & 7 & 1433 \\
    Citeseer & 3327 & 4732 & 6 & 3703 \\
    PubMed & 19717 & 44338 & 3 & 500 \\
    \hline
  \end{tabular}
\end{table}
\subsection{Baselines}
We compare MT-GCN to following state-of-the-art methods:
\begin{itemize}
  \item {\bf LP:} Label propagation algorithm used ParWalks~\cite{wu2012learning}. Partially absorbing random walk is a second-order Markov chain with partial absorption at each state.
  \item{\bf Chebyshev:} Chebyshev approach~\cite{defferrard2016convolutional} used $K$-th order Chebyshev filter to perform convolutions. The parameter $K$ is set to 2~\cite{kipf2016semi}.
  \item {\bf GCN:} GCN~\cite{kipf2016semi} follows a recursive average neighborhood aggregation scheme by stacking two graph convolutional layers. Results of GCN with validation (GCN+V) and GCN without validation (GCN-V) are considered for comparing.
  \item {\bf Co-training:} By employing a pretext random walk model to explore the global structure of the graph, GCN finds high-confidence vertices of a random walk model and adds them to the label set to train a GCN~\cite{li2018deeper}.
  \item {\bf Self-training:} Different form Co-training, Self-training trains GCN at first and secondly selects its high-confidence predictions to expand the label set. Self-training continues to train GCN with the expanded label set.
  \item {\bf Union:} Union~\cite{li2018deeper} expands the label set by integrating high-confidence predictions found by Co-training and Self-training, and continues to train the network pre-trained by Self-training with the expanded label set.
  \item {\bf Intersection:} The difference between Union and Intersection~\cite{li2018deeper} is that Intersection expands the label set by adding high-confidence predictions found by both co-training and self-training.
  \item {\bf MultiStage:} This method~\cite{sun2019multi} performs multi-stage training. At each stage, it adds high-confidence vertices with predicted pseudo labels to expand the labeled set.
  \item {\bf M3S:} Compared to MultiStage, M3S~\cite{sun2019multi} additionally utilizes DeepCluster~\cite{caron2018deep} to choose nodes with precise pseudo labels.
\end{itemize}

Since \cite{li2018deeper} and \cite{sun2019multi} used very few labeled samples, we mainly compare to their algorithms and their accuracy metrics in our Tables are from their papers, respectively.

The main difference between these comparisons and MT-GCN is that we use dual models to improve the confidence of the pseudo labels.
\subsection{Setting}
For experimental setting of MT-GCN, we use a learning rate 0.01, a dropout rate of 0.5, $\ell_2$-norm weight decay $5\times 10^{-4}$, 16 hidden units without a validation set for fair comparison, and the number of train epochs is $N=400$. After the top 200 epoches, we set $t$ to 72, 216, and 975 for Cora, Citeseer, and PubMed as in~\cite{li2018deeper}, respectively. Following \cite{li2018deeper}, we report the mean classification accuracy of 30 times on Cora and Citeseer and we average over 10 times for PubMed.
\subsection{Results}
With a low labeled rate, MT-GCN propagates label information to the entire graph efficiently. By adding pseudo labels and using the mutual teaching strategy,  MT-GCN is verified by comparing mainly with~\cite{li2018deeper} and \cite{sun2019multi}. Compared to theirs, our MT-GCN method makes full use of mutual knowledge during the training process and obtains better results than them.
\begin{table}[!ht]
\caption{Accuracy comparisons between the proposed MT-GCN and other state-of-the-art algorithms on Cora.}
\label{cora}
\centering
\begin{tabular}{l|cccc}
\hline
Labeled per class&2     & 4     & 8     & 12    \\
Label rate      &0.5\%  & 1\%   & 2\%   & 3\%   \\
\hline
LP              &56.4   &62.3   &65.4   &67.5   \\
Chebyshev       &38.0   &52.0   &62.4   &70.8   \\
GCN-V           &42.6   &56.9   &67.8   &74.9   \\
GCN+V           &50.9   &62.3   &72.2   &76.5   \\
Co-training     &56.6   &66.4   &73.5   &75.9   \\
Self-training   &53.7   &66.1   &73.8   &77.2   \\
Union           &58.5   &69.9   &75.9   &78.5   \\
Intersection    &49.7   &65.0   &72.9   &77.1   \\
MultiStage      &61.1   &63.7   &74.4   &76.1   \\
M3S             &61.5   &67.2   &75.6   &77.8   \\
\bf MT-GCN      &66.9   &73.1   &76.8   &78.5   \\
\hline
\end{tabular}
\end{table}
\begin{table}[!ht]
\caption{Accuracy comparisons between the proposed MT-GCN and other state-of-the-art algorithms on Citeseer.}
\label{citeseer}
\centering
\begin{tabular}{l|cccc}
\hline
Labeled per class&3     & 6     & 12    & 18    \\
Label rate      &0.5\%  & 1\%   & 2\%   & 3\%   \\
\hline
LP              &34.8   &40.2   &43.6   &45.3   \\
Chebyshev       &31.7   &42.8   &59.9   &66.2   \\
GCN-V           &33.4   &46.5   &62.6   &66.9   \\
GCN+V           &43.6   &55.3   &64.9   &67.5   \\
Co-training     &47.3   &55.7   &62.1   &62.5   \\
Self-training   &43.3   &58.1   &68.2   &69.8   \\
Union           &46.3   &59.1   &66.7   &66.7   \\
Intersection    &42.9   &59.1   &68.6   &70.1   \\
MultiStage      &53.0   &57.8   &63.8   &68.0   \\
M3S             &56.1   &62.1   &66.4   &70.3   \\
\bf MT-GCN      &67.7   &68.9   &69.1   &69.8   \\
\hline
\end{tabular}
\end{table}
\begin{table}[!ht]
\caption{Accuracy comparisons between the proposed MT-GCN and other state-of-the-art algorithms on PubMed.}
\label{pubmed}
\centering
\begin{tabular}{l|ccc}
\hline
Labeled per class&2   & 3     & 7     \\
Label rate      &0.03\% &0.05\% &0.1\% \\
\hline
LP              &61.4   &66.4   &65.4\\
Chebyshev       &40.4   &47.3   &51.2\\
GCN-V           &46.4   &49.7   &56.3\\
GCN+V           &60.5   &57.5   &65.9\\
Co-training     &62.2   &68.3   &72.7\\
Self-training   &51.9   &58.7   &66.8\\
Union           &58.4   &64.0   &70.7\\
Intersection    &52.0   &59.3   &69.4\\
MultiStage      &57.4   &64.3   &70.2\\
M3S             &59.2   &64.4   &70.6\\
\bf MT-GCN      &65.5   &69.5   &73.1\\
\hline
\end{tabular}
\end{table}
\begin{table}[!ht]
\caption{Accuracy comparisons between the proposed MT-GCN and other state-of-the-art algorithms under 20 labels per Class.}
\label{twenty}
\centering
\begin{tabular}{l|ccc}
\hline
Dataset         &Cora       &Citeseer   &PubMed \\
\hline
ManiReg         &59.5       &60.1       &70.7   \\
SemiEmb         &59.0       &59.6       &71.7   \\
LP              &68.0       &45.3       &63.0   \\
DeepWalk        &67.2       &43.2       &65.3   \\
ICA             &75.1       &69.1       &73.9   \\
Planetoid       &75.7       &64.7       &77.2   \\
GCN-V           &80.0       &68.1       &78.2   \\
GCN+V           &80.3       &68.9       &79.1   \\
Co-training     &79.6       &64.0       &77.1   \\
Self-training   &80.2       &67.8       &76.9   \\
Union           &80.5       &65.7       &78.3   \\
Intersection    &79.8       &69.9       &77.0   \\
\bf MT-GCN      &80.9       &69.8       &79.5   \\
\hline
\end{tabular}
\end{table}

{\bf Cora:} Table \ref{cora} reports the mean classification accuracy. As we can see, MT-GCN performs very well and outperforms most other methods by a large margin, especially with a lower label rate. As shown in Table \ref{cora}, we can see that our method achieves the best performance and outperform other baselines by a large margin.  For instance, with label rate 0.5\%, 1\%, and 2\% MT-GCN improves M3S by 5.4\%, 5.9\%, and 1.2\%, respectively.

{\bf Citeseer:} Results of Citeseer are shown in Table \ref{citeseer}. It can be seen from Table \ref{citeseer} that Union obtains the best results among these baselines. For example, with label rate 0.5\%, 1\%, and 2\%, our method improves Union by 21.4\%, 9.8\%, and 3.1\%, respectively, demonstrating the superiority of our method.

{\bf PubMed:} We report the result on PubMed in Table \ref{pubmed}. We can see that our method achieves the best performance with different label rates. Again, our methods are far better than others with lower label rates. With the label rate 0.03\% and 0.05\%, the proposed method improve M3S by 6.3\% and 3.5\%, respectively.
\subsection{Comparison of 20 labeled samples per class}
Since most of the GCN-based SSL algorithms use 20 labeled samples per class, we compare MT-GCN with the other state-of-the-art methods in Table~\ref{twenty}. The experimental setup is that we sample 20 labels for each class in all three datasets. The results of these baselines copied from \cite{kipf2016semi} and \cite{li2018deeper}. Besides the above methods, DeepWalk~\cite{perozzi2014deepwalk}, manifold regularization (ManiReg)~\cite{belkin2006manifold}, semi-supervised embedding (SemiEmb)~\cite{weston2012deep}, iterative classification algorithm (ICA)~\cite{sen2008collective}, Planetoid~\cite{yang2016revisiting} are also included. It can be seen from Table~\ref{twenty} that MT-GCN achieves state-of-the-art performance.

With the number of labeled data increases, the performance gap between the proposed MT-GCN method and other variants of GCN becomes small. It implies that the given labeled data is becoming sufficient for training a good GCN model.
\section{Conclusions}\label{Conclusions}
We propose a new strategy to train SSL GCN models with very few labeled samples, and it can enhance classification accuracy for most SSL algorithms. We train dual models with labeled samples at the beginning then pseudo labels are used for mutual teaching. Besides a supervised loss, two other loss functions are designed to update networks. A network produces pseudo labels and the other network uses the pseudo labels produced by its peer network. With the two loss functions, one network is updated with the expanded label set from its peer network.

With very few labeled samples, we obtain higher metrics than other state-of-the-art methods. Different from MT-GCN, most GCN-based methods only train network with labeled samples, which may result in the network fits unlabeled data and classification performance degrades. Different from them, we present a simple but effective graph-based SSL method, MT-GCN, which trains GCNs under extreme a low label rate, i.e., very low labeled samples per class. The idea behind MT-GCN is to maintain two GCNs simultaneously and exploits mutual knowledge between them. The mutual teaching process is accomplished by selecting the top $t$ pseudo labels for each class and adding them to enlarge the labeled data set. Experimental results on three popular datasets demonstrate the effectiveness of our method when given very few labeled data.

In the future, we will extend the strategy of mutual teaching to other domains such as image classification, sentence classification, few-shot learning, and so on. Contrastive learning can be combined with the mutual teaching strategy since contrastive learning supervised by the consistency loss in different inputs. In ML-GCN, Since the different initialization of the two layers and the dropout, predictions of the two networks are different.
\section{Acknowledgment}
This work has been supported by the National Science Foundation of China under the Grant No. 61201422.
\bibliographystyle{named}

\end{document}